\documentclass[journal=iecred,manuscript=article]{achemso}

\SectionNumbersOn
\SectionsOn

\usepackage[T1]{fontenc}       

\usepackage{graphicx}      
\usepackage{epstopdf}
\usepackage{amsmath}
\usepackage{amssymb}
\usepackage{multicol,color}
\usepackage{url}
\usepackage{amssymb}
\usepackage{framed}
\usepackage{array}
\usepackage{enumitem}
\usepackage{amsmath}
\usepackage{amsfonts}
\usepackage{amssymb}
\usepackage{wrapfig,lipsum,booktabs}
\usepackage{graphicx}
\usepackage{times}
\usepackage{t1enc}
\usepackage{amsmath,amsfonts,graphicx}
\usepackage{float}
\usepackage{epstopdf}
\usepackage{caption}
\usepackage{graphicx}
\usepackage{subcaption}
\usepackage{tikz}
\usepackage{soul}
\usepackage{enumitem}
\usetikzlibrary{shapes,arrows}
\usetikzlibrary{positioning}
\usetikzlibrary{arrows}
\usepackage{amsmath}
\tikzset{%
	block/.style    = {draw, thick, rectangle, minimum height = 3em,
		minimum width = 3em},
	sum/.style      = {draw, circle, node distance = 3cm}, 
	input/.style    = {coordinate}, 
	output/.style   = {coordinate} 
}

\author{Deepak Maurya}\email{ee11b109@ee.iitm.ac.in}
\affiliation{Department of Computer Science,  Indian Institute of Technology Madras, Chennai, India}
\altaffiliation{Robert Bosch Centre for Data Science and Artificial Intelligence}

\author{Sivadurgaprasad Chinta}
\email{sivadurgaprasad104@gmail.com}
\affiliation{Department of Chemical Engineering, Indian Institute of Technology Madras, Chennai, India}
\altaffiliation{Robert Bosch Centre for Data Science and Artificial Intelligence}

\author{Abhishek Sivaram}
\email{abhishek.sivaram21@gmail.com}

\affiliation{Department of Chemical Engineering, Indian Institute of Technology Madras, Chennai, India}

\author{Raghunathan Rengaswamy}
\email{raghur@iitm.ac.in}

\affiliation{Department of Chemical Engineering, Indian Institute of Technology Madras, Chennai, India}

\altaffiliation{Robert Bosch Centre for Data Science and Artificial Intelligence}

\title{Incorporating prior knowledge about structural constraints in model identification}

\begin{document}


\begin{abstract}
Model identification is a crucial problem in chemical industries. In recent years, there has been increasing interest in learning data-driven models utilizing partial knowledge about the system of interest. Most techniques for model identification do not provide the freedom to incorporate any partial information such as the structure of the model. In this article, we propose model identification techniques which could leverage such partial information to produce better estimates. 
Specifically, we propose Structural Principal Component Analysis (SPCA) which improvises over existing methods like PCA by utilizing the essential structural information about the model. Most of the existing methods or closely related methods use sparsity constraints which could be computationally expensive. Our proposed method is a wise modification of PCA to utilize structural information.
The efficacy of the proposed approach is demonstrated using synthetic and  industrial case-studies.
\end{abstract}



\section{Introduction}
\label{sec:intro}

Model identification is a very important task for process automation, controller implementation in chemical process industries. These models are useful for process monitoring \citep{kruger2004improved,lee2004nonlinear}, and fault detection and diagnosis \citep{maurya2005fault, choi2005fault}. In most of these applications, linear models suffice due to linearity of the process around steady state operating conditions and ease of implementation. In chemical industries it is possible to obtain partial information about the process states. Information about a subset of model equations or sparsity of the model structure can be obtained, in the form of process flow-sheets and heuristics. In order to derive better estimates of the process model, it is desired to incorporate this useful knowledge in the model identification exercise. 

Common model identification techniques lack the freedom to incorporate partial process knowledge. Consider a linear model used to describe variability in process variables. In most modeling exercises, one norm regularization is used to incorporate sparsity in the model. However, this framework does not provided the freedom to incorporate other types of information about the process, it merely makes the model sparse. In this paper, we propose a novel approach to address the problem of prior knowledge incorporation in a Principal Component Analysis (PCA) framework with appropriate and much needed modifications for solving this problem. The predominant use of PCA has been in statistical process control but PCA has also been viewed as a model identification tool as seen in several works \cite{jolliffe2002principal, paper:rao1964, narasimhan2008model, maurya2018identification}.


PCA is a multivariate technique used primarily for projecting a data set to a lower dimensional subspace, by preserving maximum variations in the data set \cite{jolliffe2002principal}, and excluding the minimal variations characterizing them as noise. The directions of maximum variability, called principal components (PCs), are used to obtain ``useful'' variations in the data, making PCA a popular denoising technique \cite{zhang2010two,chen2011denoising}. A prevalent use of PCA can be seen for statistical process control in chemometrics literature \cite{macgregor1994multivariate,macgregor1995statistical}. The key idea of these methods is based on constructing Hottelling's $T^2$ statistic \cite{hotelling1947multivariate} and using control charts such as EWMA \cite{lowry1992multivariate}, Shewhart \cite{shewhart1931economic}  and CUSUM \cite{kresta1991multivariate}. Extension of similar approach for dynamic case has been proposed by \citet{ku1995disturbance}. In this work, we concentrate on the use of PCA and its novel extensions on an entirely different problem of model identification for static case. Our primary focus lies in developing algorithms which provides the user flexibility to incorporate the prior information known about the system.

PCA can be used to derive total least squares (TLS) solution as shown by  \citet{paper:rao1964}. The directions of minimum variability can be used as directions orthogonal to the dataset, and thus can be used to obtain a set of model equations for a linear process generating the dataset \citep{narasimhan2008model,ku1995disturbance,jolliffe2002principal}. Due to the versatile nature of PCA, there have been various extensions and variants of PCA for model identification and other applications like dimensionality reduction whose applications can be seen in various engineering disciplines. Few of the key algorithmic variants of PCA include sparse PCA \cite{paper:sparse_pcaTibshirani} , robust PCA \cite{hubert2005robpca}, maximum likelihood PCA \cite{wentzell1997maximum}, probabilistic PCA \cite{kim2003process} and network component analysis \cite{liao2003network}. There are some extensions of PCA to the dynamic case also in the context of model identification as shown by \citet{ku1995disturbance, maurya2018identification}. However, in all these extensions, it is not straightforward to incorporate prior information about the process. In this paper, we specifically focus on the problem of  static linear model identification \cite{jolliffe2002principal, narasimhan2008model}.

We discuss few of the closely related works working on similar problems with slightly different assumptions. Sparse PCA \cite{paper:sparse_pcaTibshirani}, though provides a sparse representation of the data, does not inherently incorporate the information. It is primarily used to find sparse representations of high dimensional datasets \citep{shen2013consistency,shi2016sparse}. In a similar way, there does not exist a formulation to incorporate knowledge in the form of subset of model equations governing system dynamics, in conventional methods.



Another approach working along similar lines is network component analysis (NCA) \cite{liao2003network}. NCA tries to utilize the information pertaining to network structure for model identification. Similar approaches of utilizing the prior knowledge about the system can be seen in various domains of engineering. Few of the closely related approaches are robust PCA \cite{hubert2005robpca} and its variants \cite{paper:robustpca_candes2011,paper:robustpca_wright2009,paper:de2001robust,paper:huang2012robust,paper:locantore1999robust}, and extensions of sparse PCA \cite{qi2013sparse,paper:structured_SPCAjenatton2010}.
Most of these approaches have to sacrifice the simplicity in PCA formulation to incorporate the essential system information. 



In this article, we propose algorithms for estimating the entire model using the known partial process knowledge about the system. Specifically we utilize the information of non-zero and zero entries in the constraint matrix while its estimation. As an exemplar, we use the novel PCA formulation with minimal changes to incorporate the partial information available for the system. For this purpose, PCA is coupled with variable sub--selection procedures and is reported to give better estimates of the process model. The proposed algorithm is termed as structural PCA. 


The rest of the paper is organized as follows. Section \ref{sec:foundation} describes a formal description of problem setting, assumptions and basic introduction to PCA in the context of model identification. PCA is also discussed in Appendix Section \ref{sec:pca_MI} in detail. Further, Section \ref{sec:SPCA} in the main paper describes the proposed structural PCA (sPCA) algorithm. The key idea of the sPCA algorithm is to consecutively estimate each linear relation sequentially in an independent manner. We further improvise the SPCA algorithm results in Section \ref{sec:cspca} by leveraging the information obtained from few of the already estimated linear relations. To utilize the information from few of the already estimated liner relations, we propose constraint PCA (cPCA) in Appendix Section \ref{sec:cpca}. In Section \ref{sec:cspca}, we combine cPCA and sPCA algorithm and hence name the algorithm as CSPCA algorithm. We also demonstrate the efficacy of proposed algorithms in various numerical case studies. Concluding remarks and directions to future work are discussed in Section \ref{sec:conc}.

\section{Foundations}
\label{sec:foundation}
We start the discussion on model identification problem for noise-free data. 
As seen in the literature, PCA has been predominantly used in identifying directions of maximum variability and subsequent utilization of this analysis for monitoring problems, PCA can be also viewed as one of the approaches for model identification \cite{jolliffe2002principal, paper:rao1964}.  Our intention lies in exploiting this viewpoint towards solving the problem of prior knowledge incorporation. 

Let $\mathbf{x}(t)$ be a $n \times 1$ vector consisting measurements of $n$ variables at time instant $t$. It is assumed that these $n$ variables are related by $m$ linear equations at all time instants and in this manuscript we assume $m$ is known apriori. This may be formally stated as 
\begin{align}
\mathbf{A_0}\mathbf{x}(t) = \mathbf{0}_{m \times 1} \quad \forall t 
\label{eq:noise_free_mdl}
\end{align}     
where $\mathbf{A_0} \in \mathbf{R}^{m \times n} $ is a time-invariant constraint matrix. In this paper, $\mathbf{A}$ or constraint matrix is interchangeably referred to as model.
At each time instant, measurement $\mathbf{y}(t)$ of all the $n$ variables is assumed to be corrupted by noise
\begin{align}
\mathbf{y}(t)  = \mathbf{x}(t) + \mathbf{e}(t)
\label{eq:yt_def}
\end{align}

The following assumptions are made on the random errors:
\begin{enumerate}
	\setlength\itemsep{0.1em}
	\item $\begin{aligned}[t]
	\mathbf{e}(t) \sim \mathcal{N}(\mathbf{0},\sigma^2\mathbf{I})
	\end{aligned}$
	\item $\begin{aligned}[t]
	\mathbb{E}(\mathbf{e}(j)\mathbf{e}^T(k)) = \sigma^2\delta_{jk}\mathbf{I}_{n \times n} 
	\end{aligned}$
\end{enumerate}

where $\mathbb{E}(.)$ is the usual expectation operator and  $\mathbf{e}(t)$ is a vector of white-noise errors, with all elements having identical variance $\sigma^2$ as stated above. 
We introduce the collection of $N$ such noisy measurements as follows 
\begin{align}
\mathbf{X} & = \begin{bmatrix}
\mathbf{x}[0] & \mathbf{x}[1] &  \cdots & \mathbf{x}[N-1] \end{bmatrix} \\
\mathbf{Y} & = \begin{bmatrix}
\mathbf{y}[0] & \mathbf{y}[1] &  \cdots & \mathbf{y}[N-1] \end{bmatrix}
\label{eq:N_measr}
\end{align}
Given $N$ noisy measurements of $n$ variables, the objective of PCA algorithm is to estimate the constraint model $\mathbf{A}_0$ in \eqref{eq:noise_free_mdl}. 
We formally describe theoretically relevant aspects of PCA in Appendix \ref{sec:pca_MI} and focus on problem of our interest in the next section.


\section{Model Identification with known model structure (sPCA)}
\label{sec:SPCA}
In this section, we describe the main challenging and practical problem of incorporating the knowledge about structure of the entire constraint matrix during its estimation. This essentially means we assume to have a priori knowledge about the set of variables which satisfy each linear relationship. For example, the structure of constraint matrix for flow-mixing case study presented in Figure \ref{fig:fl-mix} would be 
\begin{align}
\mathbf{  structure(A_0)} &= \begin{bmatrix}
\times    & \times &    0 &    0 &    \times\\
0    &  \times &      \times &     0&     0\\
0    & 0 &   \times &     \times &    \times \\
\end{bmatrix} 
\label{eq:flmix_Astruct}
\end{align}
The above structure provides us the essential information about the set of variables combining linearly at each node of flow network. This information about which variables are related by linear relation may be easily available in flow distribution networks \cite{narasimhan2008model}. Utilizing this valuable information in the formulation of optimization problem (one optimization problem for each constraint) for estimation of constraint matrix will lead us to a better solution. 

In this section, we present a novel approach to estimate the constraint matrix of a given structure without getting drowned into imposing sparsity constraints. The key difference in the methodology of the proposed algorithm and the existing frameworks is to estimate each row of the constraint matrix, meaning each linear relation separately rather than the whole constraint matrix. The linear relations estimated sequentially are stacked together at the end to construct the entire constraint matrix. 

This idea of estimating linear relations separately equips us with considerable freedom to incorporate the structural constraints without diving into sparsity constraints which can be computationally expensive. Our proposed approach utilizes wisely modified version of PCA to estimate the constraint matrix. This brings in some new challenges which are addressed in a detailed manner. In order to demonstrate wide range of challenges and the proposed remedies, few simple constraint matrices are considered. We first consider a simple example to demonstrate the key idea of sPCA algorithm and the improvement it provides over PCA.  

The key idea of sPCA algorithm is estimating linear relations corresponding to each row of constraint matrix structure separately via sub-selection of variables. For example, consider, consider a simple flow mixing network example shown in Figure \ref{fig:fl-mix}:  
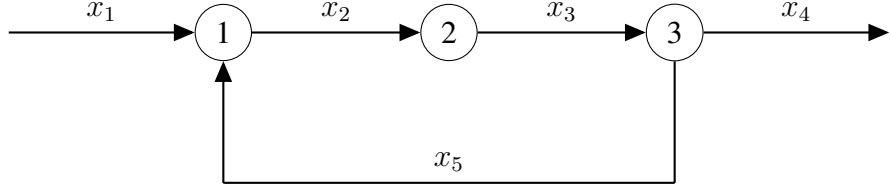
\begin{figure}[H]
	\centering
	\begin{tikzpicture}
	
	\tikzset{vertex/.style = {shape=circle,draw,minimum size=1.5em}}
	\tikzset{edge/.style = {draw, -latex',thick,->,>= triangle 45}}
	\node (1p) at (-3,0) {};		
	\node (3n) at (9,0){};
	\node [vertex](1) at (0, 0) {1};
	\node [vertex](2) at (3,0) {2};
	\node [vertex](3) at (6, 0) {3};

	\path [edge] (1p)--(1) node [draw = none,midway,above]{$x_1$} ;
	\path [edge]  (1)--(2) node [draw = none,midway,above]{$x_2$};
	\path [edge] (2)--(3) node [draw = none,midway,above]{$x_3$};
	\path [edge] (3)--(3n) node [draw = none,midway,above]{$x_4$};
	\path[edge] (3) -- (6,-2)  (6,-2)-- (0,-2) node [draw = none,midway,above]{$x_5$}  (0,-2)--(1);
	
	\end{tikzpicture}
	\caption{Flow mixing case study} \label{fig:fl-mix}
\end{figure}
The structure of constraint matrix is given below:
\begin{align}
	\mathbf{\mathbf{A_0}} &= \begin{bmatrix}
	1    & -1 &    0 &    0 &    1\\
	0    & 1&     -1&     0&     0\\
	0    & 0 &   1 &     -1&    -1 \\\end{bmatrix} \label{eq:flmix_trueA0}
\end{align}
This network could be easily seen in various engineering disciplines like electrical circuits or water distribution in pipelines. The flow balance at each node, at any time instant $t$ can be stated as 
\begin{subequations}
	\begin{align}
	x_1(t) - &x_2(t) + x_5(t) = 0, \qquad \text{Node 1} \\
	x_2(t) - &x_3(t) = 0, \qquad \qquad \hspace{0.54cm}\text{Node 2} \\
	x_3(t) - &x_4(t) - x_5(t)= 0, \qquad \text{Node 3}
	\end{align}
\end{subequations}
The model equation of this flow network corresponding to \textit{noise-free} measurements at three nodes can be stated as , $\mathbf{A_0} \mathbf{x}(t) = \mathbf{0}$, where,
	\begin{align}
	\mathbf{\mathbf{A_0}} &= \begin{bmatrix}
	1    & -1 &    0 &    0 &    1\\
	0    & 1&     -1&     0&     0\\
	0    & 0 &   1 &     -1&    -1 \\\end{bmatrix} \tag{\ref{eq:flmix_trueA0}}\\
	\mathbf{x}(t) &= \begin{bmatrix}
	x_1(t) & x_2(t) & x_3(t) & x_4(t) & x_5(t)\\
	\end{bmatrix}^\top
	\end{align}

We further discuss the process to generate synthetic data corresponding to the above system. The \textit{noise-free} measurements are generated by utilizing the null space of constraint matrix $\mathbf{A_0}$. For any general matrix $\mathbf{ A_0} \in \mathbb{R}^{m \times n}$, the null space denoted by $\mathbf{  A_{0}^{\perp}}$ follows:
\begin{align}
\mathbf{ A_0 \mathbf{  A_{0}^{\perp}}} = \mathbf{ 0}_{m \times (n-m)}, \qquad \text{where} \quad \mathbf{ \mathbf{  A_{0}^{\perp}}} \in \mathbb{R}^{n \times (n-m)}, \qquad \text{rank}(\mathbf{ A_0}) =  m < n
\end{align}
Given a model $\mathbf{\mathbf{A_0}\mathbf{X} = 0}$, it can be seen that $\mathbf{\mathbf{X}}$ lies in the null-space of $\mathbf{\mathbf{\mathbf{A_0}}}$. Hence, the data is generated by using the null space of $\mathbf{\mathbf{\mathbf{A_0}}}$, and obtaining $\mathbf{\mathbf{X}}$ by a linear combination of the null-space with random numbers. It could be formally stated as 
\begin{align}
\mathbf{ X} = \mathbf{  \mathbf{  A_{0}^{\perp}} M}, \qquad  \mathbf{ M} \in \mathbb{R}^{(n-m) \times N}
\label{eq:dgp_noisefree}
\end{align}
where, $\mathbf{ M}$ contains the random coefficients. It could be easily verified $\mathbf{ A_0 X} = \mathbf{ 0}$ from \eqref{eq:dgp_noisefree}.

As stated in section \ref{sec:foundation}, the \textit{noise-free} measurements -- $\mathbf{ x}(t)$ are not accessible. Instead, we are supplied the noisy measurements of $\mathbf{ x}(t)$, denoted by $\mathbf{ y}(t)$ in \eqref{eq:yt_def}. It is assumed that a collection of $N$ such noisy measurements are available as stated in \eqref{eq:N_measr}. The noise used to corrupt the true measurements is white Gaussian noise with a signal to noise (SNR) ratio as 10. SNR is formally defined as the ratio of variance of noise-free signal to the variance of its noise. 

The constraint matrix can be estimated can be estimated by applying PCA to the subset of variables participating at each node separately. For instance at node 1 in Figure \ref{fig:fl-mix}, variables $y_1$, $y_2$ and $y_5$ will be considered.
\begin{align}
    \mathbf{y_{sub1}(t)} = \begin{bmatrix}
    y_1(t) & y_2(t) & y_5(t)
    \end{bmatrix}
    \label{eq:fl-mix_subsel1}
\end{align}
Applying PCA on a collection of $N$ measurements of $\mathbf{y_{sub1}(t)}$ will deliver us a row vector $\mathbf{a_{sub1}}$ of dimension $1 \times 3$ such that $\mathbf{a_{sub1} x^T_{sub1}(t)} = 0$, where $\mathbf{x_{sub1}(t)}$ contains the noise-free measurements of sub-selected set of variables commensurate to $\mathbf{y_{sub1}(t)}$ in \eqref{eq:fl-mix_subsel1}. It should be noted that estimated constraint row vector will only contain the non-zero entries corresponding to sub-selected variables. Basically, we mean that the structure will be,
\begin{align}
    \mathbf{\hat{a}_{sub1}} = \begin{bmatrix}
    \hat{a}_{11} & \hat{a}_{21} & \hat{a}_{51}
    \end{bmatrix}
    \label{eq:fl-mix_Asub1form}
\end{align}
where, $a_{i1}$ correspond to the coefficient of $i^\text{th}$ variable. The desired structure for the first row of the constraint matrix could be constructed by appending zeros at the desired locations as shown below
\begin{align}
    \mathbf{\hat{a}_1} = \begin{bmatrix}
    \hat{a}_{11} & \hat{a}_{21} & 0 & 0 & \hat{a}_{51}
    \end{bmatrix}
\end{align}
This procedure could be similarly applied at nodes $2$ and $3$ in Figure \ref{fig:fl-mix} to estimate row constraint vectors $\mathbf{\hat{a}_2}$ and $\mathbf{\hat{a}_3}$ respectively. The  entire constraint matrix can be constructed by stacking the estimated linear relations
\begin{align}
    \mathbf{\hat{A}_{spca}} = \begin{bmatrix}
    \mathbf{\hat{a}_1} \\
    \mathbf{\hat{a}_2} \\
    \mathbf{\hat{a}_3}
    \end{bmatrix}
    \label{eq:spca_Aflmix_recon}
\end{align}
To investigate the goodness of estimates, we utilize the subspace-dependence based metric  stated in \citet{narasimhan2008model} and briefly mentioned here. The subspace-dependence metric can be viewed as distance between the row spaces of the true ($\mathbf{  A_0}$) and estimated constraint matrix ($\mathbf{  \hat{ A}}$). The minimum distance of each row of  $\mathbf{  A_0}$ from the row space of  $\mathbf{  \hat{ A}}$ in least squares sense is given by
\begin{align}
	\theta = \sum_{i=1}^{m} || \mathbf{  A_0}_i - \mathbf{  A_0}_i \mathbf{  \hat{ A}^\top }\left( \mathbf{ \hat{ A} \hat{ A}^\top}\right)^{-1} \mathbf{  \hat{ A} }  ||
	\label{eq:subspace_metr}
\end{align}
where subscript $i$ in $\mathbf{A_0}_i$ denotes  $i^\text{th}$ row. 

The subspace dependence metric mentioned in \eqref{eq:subspace_metr} is used for the evaluation of efficacy of estimated constraint matrix by the proposed algorithm, which we term as structural principal component analysis (sPCA). The following numbers are reported for $1000$ runs of MC simulations with SNR = 10.
\begin{align}
 \theta_{PCA} = 0.1293, \qquad \qquad  \theta_{sPCA} = 0.1188
    \label{eq:fl-mix_spcaest}
\end{align}
It can be easily inferred from equation \eqref{eq:fl-mix_spcaest} that sPCA estimate is much closer to the true constraint matrix compared to PCA. 

Consider the flow balance across the all the nodes 1,2 and 3 as shown in the figure below 
\begin{figure}[H]
	\centering
	\begin{tikzpicture}
	
	
	\tikzset{vertex/.style = {shape=circle,draw,minimum size=1.5em}}
	\tikzset{edge/.style = {draw, -latex',thick,->,>= triangle 45}}
	\node (1p) at (-3,0) {};		
	\node (3n) at (9,0){};
	\node [vertex](1) at (0, 0) {1};
	\node [vertex](2) at (3,0) {2};
	\node [vertex](3) at (6, 0) {3};

	\path [edge] (1p)--(1) node [draw = none,midway,above]{$x_1$} ;
	\path [edge]  (1)--(2) node [draw = none,midway,above]{$x_2$};
	\path [edge] (2)--(3) node [draw = none,midway,above]{$x_3$};
	\path [edge] (3)--(3n) node [draw = none,midway,above]{$x_4$};
	\path[edge] (3) -- (6,-2)  (6,-2)-- (0,-2) node [draw = none,midway,above]{$x_5$}  (0,-2)--(1);
	
	\draw [color=red,thick](-1.1,-2.3) rectangle (6.9,1);
	\end{tikzpicture}
	\caption{Flow mixing case study} \label{fig:fl-mix_boun}
\end{figure}
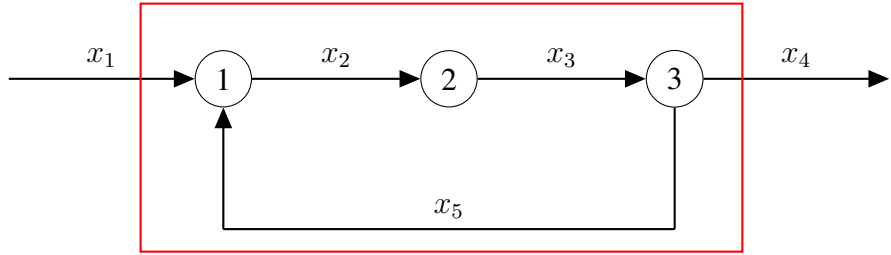
It basically shows that node $1, 2$ and $3$ can be considered as a single node to derive the linear relation among variable $x_1$ and $x_4$. So applying traditional PCA  may reveal the linear relation among the variables $x_1$ and $x_4$.  The corresponding equivalent structure of the constraint matrix is stated below: 
\begin{align}
\text{structure}(\mathbf{  A_0}) &= \begin{bmatrix}
\times    & \times &    0 &    0 &    \times\\
0    &  \times &      \times &     0&     0\\
\times   & 0 &   0 &     \times &    0 \\
\end{bmatrix} 
\label{eq:flmix_Astruct_equvl}
\end{align}
Hence, only the constraint corresponding to last row in Eq \ref{eq:flmix_Astruct} and Eq \ref{eq:flmix_Astruct_equvl} is different. 

Unfortunately this phenomena creates a challenging issue which can be dealt with appropriate modification in the sPCA approach discussed previously. To illustrate this phenomena, let us consider another simple example of desired constraint matrix  stated below:
\begin{align}
\text{structure}(\mathbf{  A_0}) &= \begin{bmatrix}
\times    & \times &    \times &  \times &  0 &    \times\\
\times    &  \times &      \times& \times &     0&     0\\
\times    &  \times  &      \times &    0 & 0&     0\\
\times    & \times &   \times &  \times &   0 &    0 \\
\end{bmatrix} 
\label{eq:spca_Astruct_runex}
\end{align}
We intend to estimate each linear relation separately starting from the first row of $\mathbf{  structure(A_0)} $ specified in equation \eqref{eq:spca_Astruct_runex}. The sub-selected variables would be 
\begin{align}
	\mathbf{y_{sub1}(t)} = \begin{bmatrix}
	y_1(t) & y_2(t) & y_3(t) & y_4(t) & y_6(t) 
	\end{bmatrix}
	\label{eq:spcarunex_subsel1}
\end{align}
Applying PCA on $N$ measurements of $\mathbf{y_{sub1}(t)} $  may not deliver us the desired structure specified in the first row of $\mathbf{  structure(A_0)}$ in \eqref{eq:spca_Astruct_runex}. This may occur as the complementary set of zero locations in row 2, 3 and 4 of $\mathbf{  structure(A_0)}$ in \eqref{eq:spca_Astruct_runex} are a subset of the complementary set of zero locations in row 1. It basically means the idea of applying PCA on sub-selected variables doesn't gaurentee  non-zero coefficient of the selected variables.  Sub-selection only gaurentees the zero coefficient of the discarded variables. Ignoring this fact could lead us to estimate a linear relation corresponding to structure specified in row 2 of \eqref{eq:spca_Astruct_runex} when we intend to estimate the relation corresponding to structure of row 1. If we ignore the above scenario and proceed to estimate $2^\text{nd}$ row of the constraint matrix with the desired structure by sub-selection of variables, we may end up in estimating same previously estimated linear relation. This may also lead us to miss out the first constraint as the variable $x_6(t)$ will not be sub-selected in any of the consecutive iterations.  


We propose a novel approach to deal with such scenario. The primary concern was ambiguity in the estimated relationship to be of the structure we intended. This issue raises doubts mainly due to estimation of constraint with more zero entries afterwards. Such a case could be avoided by re-configuring the structure of given constraint matrix. As we intend to estimate the constraint with less number of zeros afterwards, corresponding rows are pushed down.  So, the constraint matrix is re-structured in ascending order of the number of non-zero locations in each row. The objective of this step is to avoid estimation of the individual constraints which are already estimated. The re-structured constraint matrix for \eqref{eq:spca_Astruct_runex} can be stated as 
\begin{align}
\text{re-structured}(\mathbf{  A_0}) &= \begin{bmatrix}
\times    &  \times  &      \times & 0 &    0&     0\\
\times    &  \times &      \times&      \times &     0&     0\\
\times    & \times &   \times &      \times&     0 &    0 \\
\times    & \times &    \times &      \times&    0 &    \times\\
\end{bmatrix} 
\label{eq:spca_Arestruct_runex}
\end{align} 

This rewarding step ensures obtaining the constraint with lower cardinality of non-zero elements before compared to constraints with higher cardinality. But, it still does not resolve the ambiguity in obtaining same constraints with similar structure.
We propose a two-step remedy which is illustrated as follows:
\begin{enumerate}
	\item \label{item:detect} \textit{ Detection}: Such cases could be identified by a rank check of the linear relation obtained at each step. Let the constraint matrix up to $i^\text{th}$ row be $\mathbf{ \hat{ A}}_i$ and the linear relation obtained from $(i+1)^\text{th}$ row be $\mathbf{  \hat{ a}}_{i+1} $. If we obtain a constraint at $(i+1)^\text{th}$ step which is just a linear combination of previously estimated constraints, then rank of $\left( \begin{bmatrix}
	\mathbf{ \hat{ A}}_i \\
	\mathbf{  \hat{ a}}_{i+1}
	\end{bmatrix}\right)$ will be the same as rank of $\mathbf{ \hat{ A}}_i$. This idea is used for detection of previously estimated constraint. 
	\item \textit{Identification}: It should be noted that the cause for detecting a previously estimated constraint is existence of multiple constraints. 
	
	In order to filter the right constraint from a set of multiple constraints, the idea of rank check is utilized again. Let the full row rank constraint matrix estimated up to $i^\text{th}$ row be $\mathbf{  \hat{ A}}_i$. For $(i+1)^\text{th}$ row, we propose to consider all the eigenvectors instead of one eigenvector corresponding to minimum eigenvalue. This is done because the set of all eigenvectors is a superset of all the constraints identified till $(i+1)^\text{th}$ iteration. 
	
	For example in $2^\text{nd}$ iteration for the structure provided in \eqref{eq:spca_Arestruct_runex},  the subset of variables would be  
	\begin{align}
	\mathbf{y_{sub2}(t)} = \begin{bmatrix}
	y_1(t) & y_2(t) & y_3(t)  & y_4(t)
	\end{bmatrix}
	\label{eq:spcarunex_subsel2_a}
	\end{align}
	Applying PCA on $N$ measurements of $\mathbf{y_{sub2}(t)}$ should ideally reveal $3$ linear relations. But it is known to us from the given structure that there exist only $2$ linear constraints for this particular row-structure. Those $2$ linear relations can be filtered from the $3$ constraints using  rank check. The above procedure is formally stated below. 
	
	 We define the matrix $\mathbf{\hat{ B}}_{i+1}$ which contain the eigenvectors along its rows in $(i+1)^\text{th}$ iteration. It should be noted that these eigenvectors are arranged along the rows such that the eigenvalues are increasing with increasing row  numbers. 
	 Let the dimension of $\mathbf{\hat{ B}}_{i+1}$ be $n_{i+1} \times n_{i+1}$ and its $j^\text{th}$ row be denoted by $\mathbf{\hat{b}}_{i+1,j}$.  
	 
	 First, we make the hypothesis that the $j^\text{th}$ row of $\mathbf{\hat{ B}}_{i+1}$ - $\mathbf{\hat{b}}_{i+1,j}$ contains a constraint. We define 
	 \begin{align}
	 	\mathbf{  \hat{ A}_{i,j}} = \left( \begin{bmatrix}
	 	\mathbf{  \hat{ A}}_i \\
	 	\mathbf{\hat{b}}_{i+1,j}
	 	\end{bmatrix} \right)
	 \end{align}
	 To test this hypothesis, we compare the rank of $\mathbf{  \hat{ A}_{i,j}}$ and $\mathbf{  \hat{ A}}_i$. If the ranks of both matrices are equal, then $\mathbf{\hat{b}}_{i+1,j}$ is rejected, otherwise $\mathbf{  \hat{ A}}_i$ is updated using 
	 \begin{align}
	 \mathbf{  \hat{ A}}_i = \left( \begin{bmatrix}
	 \mathbf{  \hat{ A}}_i \\
	 \mathbf{\hat{b}}_{i+1,j}
	 \end{bmatrix} \right)
	 \end{align}
	 because it contains a new relation. 
	 
	 The number of constraints to be chosen from this $(i+1)^\text{th}$ iteration will be known from the given structure. Let it be $m_{i+1}$. So this process of detection and filtering right constraint is carried out until $m_{i+1}$ constraints are identified.
\end{enumerate}

The estimated constraint matrix could be easily reconfigured according to original specified structure once all the constraints are estimated for the re-structured $\mathbf{  A_0}$.

In this section, we discussed the main theme of sub-selecting variables in the proposed algorithm with the help of flow-mixing case-study. This example demonstrated the efficacy of the results via proposed algorithm. Later on, various challenges and remedies will be illustrated with the help of another constraint matrix. We close this section by presenting the full and final version of proposed algorithm in Table \ref{tab:SPCA_algo}. Three diverse case-studies are presented in the next sub-section to show the utility and performance of the proposed algorithm.

\begin{table}[H]
	\caption{Structured PCA (sPCA) Algorithm }
	\label{tab:SPCA_algo}
	\hrulefill
	\begin{enumerate}
		\item \label{item:reconA} Given the structure of constraint matrix $\mathbf{  A}_{\text{struct}}$ of dimension $m \times n$ configure it  such that
		\begin{align}
			f(i+1) \geq f(i) \quad \forall \quad i \in \{1,2,..,m-1\}
		\end{align} 
		where $f(i): \text{number of non-zero elements in row } i$ of $\mathbf{  A}_{\text{struct}}$. Let the re-configured matrix be $\mathbf{  A}_{\text{re-struct}}$. Let $g(j)$ be the count of number of rows in $\mathbf{  A}_{\text{re-struct}}$  having similar structure with $j^\text{th}$ row of $\mathbf{  A}_{\text{re-struct}}$. Initialize $\mathbf{  \hat{ A}}_{\text{est},i} = \begin{bmatrix}
		& 
		\end{bmatrix}$ for iteration $i = 1$.
		\item \label{item:struct-sim} For iteration $i > 2$, perform the structure similarity test of $i^\text{th}$ and $j^\text{th}$ rows of   $\mathbf{  A}_{\text{re-struct}}$, where $j \in \{1,2,...,(i-1)\}$. If there is any match, discard the $i^\text{th}$ row of $\mathbf{  A}_{\text{re-struct}}$ and revisit step \ref{item:struct-sim} with $i = i + 1$. If there is no match, proceed to step \ref{item:sub-sel}.
		
		\item \label{item:sub-sel} For iteration $i$, apply PCA on the sub-selected set of variables from $\mathbf{  Y}$ corresponding to structure of $i^\text{th}$ row of  $\mathbf{  A}_{\text{re-struct}}$. Let the number of sub-selected variables and measurements matrix be $n_{\text{sub},i}$ and $\mathbf{  Y}_{\text{sub},i}$ respectively. Collect all eigenvectors of sample covariance matrix of $\mathbf{  Y}_{\text{sub},i}$ to obtain $\mathbf{ \hat{A}}_{\text{sub},i}$.
		\item Include zeros in  $\mathbf{ \hat{A}}_{\text{sub},i}$ corresponding to the structure of $i^\text{th}$ row in  $\mathbf{  A}_{\text{re-struct}}$ to obtain $\mathbf{ \hat{A}}_{i}$. Note that the dimension of $\mathbf{ \hat{A}}_{i}$ is $n_{\text{sub},i} \times n$. 
		\item Filter the correct linear relations by performing rank test on constraints identified in iteration $i$. For $k = \{1,2,...,n_{\text{sub},i}\}$
		{\small \begin{align}
			\mathbf{\hat{ A}}_{\text{est},i} = \begin{cases}
			\mathbf{\hat{ A}}_{\text{est},i} & \quad \text{rank}\left(\mathbf{\hat{ A}}_{\text{est},i}\right) = \text{rank}\left(\mathbf{  \hat{ A}}_{\text{est},i,k}\right), \\
			\begin{bmatrix}
			\mathbf{  \hat{ A}}_{\text{est},i} \\ 
			\mathbf{ \hat{A}}_{i}(k,:)
			\end{bmatrix} & \quad  \text{rank}\left(\mathbf{\hat{ A}}_{\text{est},i}\right) \neq \text{rank}\left(\mathbf{  \hat{ A}}_{\text{est},i,k}\right) \& \quad \text{nrow}\left(\mathbf{\hat{ A}}_{\text{est},i}\right) - \text{nrow}\left(\mathbf{\hat{ A}}_{\text{est},i-1}\right) < g(i)
			\end{cases}
		\end{align}}
		where $\mathbf{  \hat{ A}}_{\text{est},i,k} = \left(\begin{bmatrix}
		\mathbf{  \hat{ A}}_{\text{est},i} \\ 
		\mathbf{ \hat{A}}_{i}(k,:)
		\end{bmatrix}\right)$, $\mathbf{ \hat{A}}_{i}(k,:)$ denotes the $k^\text{th}$ row of  $\mathbf{ \hat{A}}_{i}$, $\text{nrow}\left(\mathbf{\hat{ A}}_{\text{est},i}\right)$ denotes the number of rows in $\mathbf{\hat{ A}}_{\text{est},i}$ and $g(i)$ is defined in step \ref{item:reconA}. This step may be terminated for a $k$ satisfying $\text{nrow}\left(\mathbf{\hat{ A}}_{\text{est},i}\right) - \text{nrow}\left(\mathbf{\hat{ A}}_{\text{est},i-1}\right) = g(i)$ in order to improve computational efficiency. 
		
		\item Repeat the entire procedure from step \ref{item:struct-sim} until $\text{nrow}(\mathbf{  \hat{ A}}_{\text{est},i+1}) < m$.
		\item Map the estimated constraint matrix to the original form supplied by user in step \ref{item:reconA}.
	\end{enumerate}
	\hrulefill
\end{table}


\subsection{Case-study 1}
\label{sec:spca_simlex1}
This is a synthesised case study to show the efficacy of proposed approach, when the structure of the constraints are known. The original constraints and the structural information of the same are given as below. Constraint matrix consists of six variables, in which two variables are out of the scope (i.e. absent) for the constraints considered in this case study. 
\begin{align}
	\mathbf{  A_0} = \begin{bmatrix}
		1  &   1   & 0   &  0  &   0  &  0 \\
		1  &   2   & 3   &  0  &   0  &   0 \\
		3  &   1   & -1   &  2  &   0  &   0 \\
	\end{bmatrix}, 
	\qquad
	\text{structure}(\mathbf{  A_0}) = \begin{bmatrix}
	\times    & \times &    0 &    0 & 0  &  0\\
	\times    &  \times &      \times &     0 &  0  &   0\\
	\times    & \times &   \times &     \times  & 0  &   0 \\
	\end{bmatrix}
	\label{eq:spca_siml_A0_1}
\end{align}

\begin{figure}[H]
   	\centering
   	\includegraphics[width=0.65\linewidth]{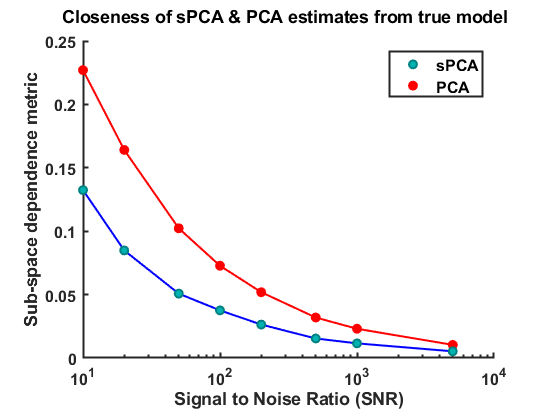}
   	\caption{Comparison of Model estimates by sPCA and PCA at different SNRs}
   	\label{fig:simul1_spcasnrtheta}
 \end{figure}

To compare the proposed sPCA approach with the traditional PCA, 500 MC simulations have been tested for SNR values 10, 20, 50, 100, 200, 500, 1000 and 5000. For each MC simulation at each SNR value, data is generated for 1000 random samples. Sub-space dependence metric is evaluated for each constraint matrix and is averaged at each SNR value. These metric values can be obtained from figure \ref{fig:simul1_spcasnrtheta}, it can be observed from the figure that including available process information can improve the estimates.

\subsection{Case-study 2 }

The system considered in this case study is steam melting network, which is considered by many researchers for testing data reconciliation and gross error detection approaches \cite{narasimhan2008model,serth1986gross,sun2011gross}. The network contains 28 flow variables and 11 flow constraints. The data is generated by varying 17 flows (F4, F6, F10, F11, F13, F14, F16 - F22, F24, F26 - F28) independently using a first order ARX model for 1000 time samples, the flow rates of remaining flows are obtained by using the flow constraints at each time sample. The flowsheet of the steam melting network can be observed below in Figure \ref{fig:methanol_case}. 

\begin{figure}[H]
   	\centering
   	\includegraphics[width=0.65\linewidth]{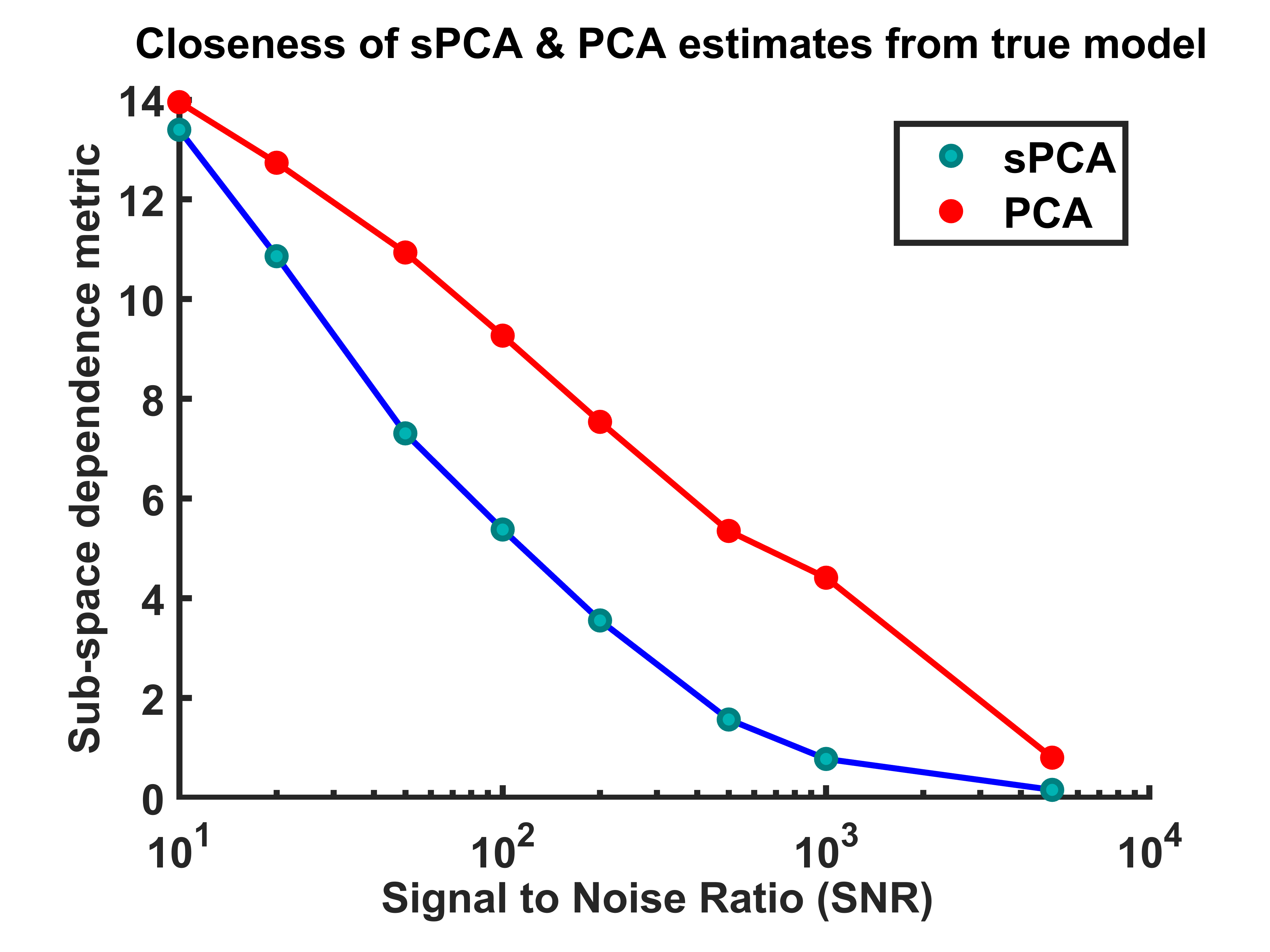}
   	\caption{Comparison of PCA and sPCA}
   	\label{fig:methanol_pcavsspca}
  \end{figure}

Assuming the structure of the plant is known, flow constraint matrix is estimated using both PCA and sPCA for 1000 runs of each SNR value. The mean closeness measure of the constructed constraint matrices to the original matrix for different SNR values are provided in Figure \ref{fig:methanol_pcavsspca}. It is interesting to note that except for SNR  10, sPCA delivers better estimates than PCA in all 1000 runs. 

\begin{figure}[H]
   	\centering
   	\includegraphics[width=0.65\linewidth]{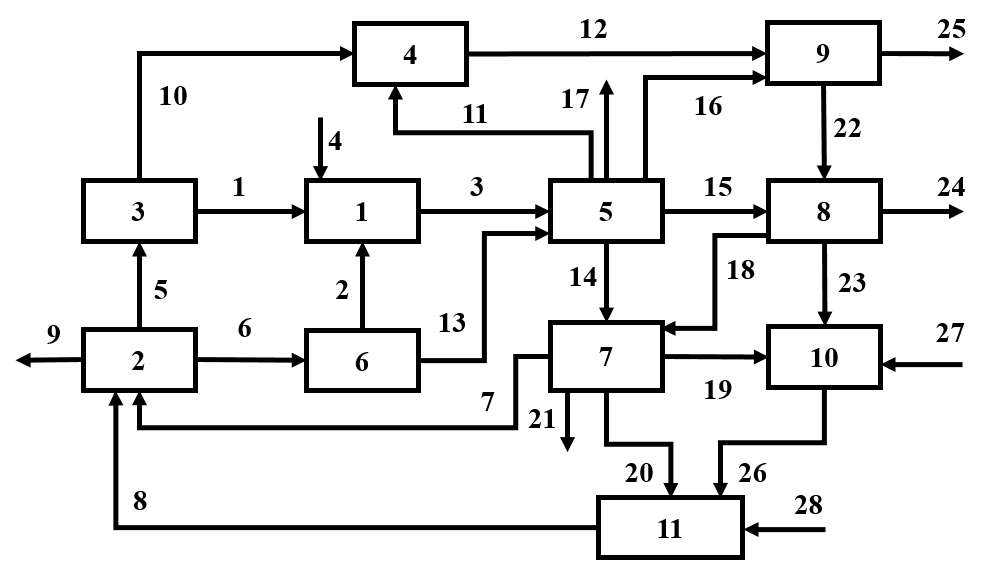}
   	\caption{Flow network of steam melting network for methanol synthesis plant}
   	\label{fig:methanol_case}
  \end{figure}

\subsection{Case-study 3}
\label{sec:spca_ex3}
We intend to show the supremacy of model estimates obtained by sPCA algorithm in this simulation-study. We consider the system with constraint model mentioned in \eqref{eq:spca_Astruct_runex}.

The model is assumed to be $\mathbf{  A_0 X = 0}$, where 
\begin{align}
	\mathbf{  A_0} = \begin{bmatrix}
		3  &   1   & -1   &  2  &   0  &  -6 \\
		2  &   1   & -2   &  1  &   0  &   0 \\
		1  &   1   & -1   &  0  &   0  &   0 \\
		1  &  -3   &  1   &  1  &   0  &   0
	\end{bmatrix}
	\label{eq:spca_siml_A0}
\end{align}
Please note that the structure of $\mathbf{  A_0} $ in \eqref{eq:spca_siml_A0} matches with structure specified in \eqref{eq:spca_Astruct_runex}. Data is generated with the same procedure followed in flow mixing case study in Figure \ref{fig:fl-mix}. 

We perform MC simulations of 100 runs at various signal to noise ratio (SNRs) to demonstrate the goodness of estimates obtained by proposed algorithm -- sPCA. For the purpose of comparison, model was estimated from PCA algorithm too and subspace dependence metric defined in \eqref{eq:subspace_metr} is used to evaluate the quality of obtained estimates. For each realization, the structure passed to sPCA algorithm is 
\begin{align}
	\text{structure}(\mathbf{  A_0}) = \begin{bmatrix}
	\times    & \times &    \times &    \times & 0  &  \times\\
	\times    &  \times &      \times &     \times &  0  &   0\\
	\times    & \times  &      \times &     0&     0&0  &\\
	\times    & \times &   \times &     \times  & 0  &   0 \\
	\end{bmatrix}
	\label{eq:spca_ex1}
\end{align}
The results from PCA and sPCA are presented below:

\begin{figure}[H]
	\centering
	\begin{minipage}{.5\textwidth}
	\centering
   	\includegraphics[width=0.95\linewidth]{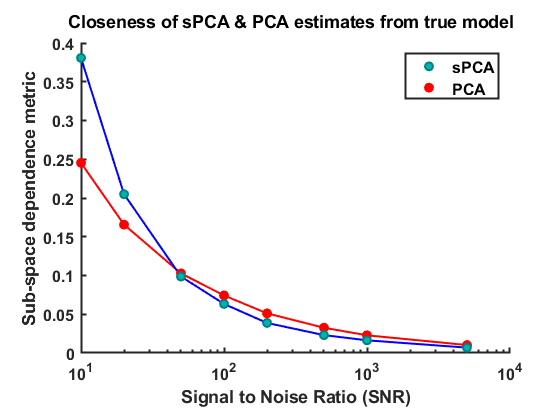}
   	\label{fig:simul2spcasnrtheta}
	\end{minipage}%
	\begin{minipage}{.5\textwidth}
		\centering
		\includegraphics[width=0.95\linewidth]{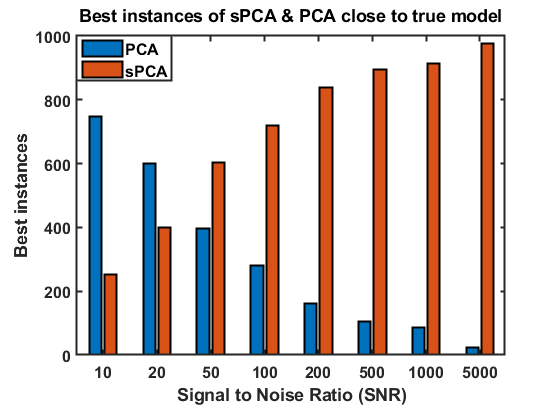}
   	\label{fig:simul2spcasnrtheta_1}
	\end{minipage}
	\caption{Comparison of Model estimates by sPCA and PCA at different SNRs}
	\label{fig:spca_cs3}
\end{figure}


From the plot, it can be easily noticed that sPCA outperforms PCA at SNR above 50. It is also interesting to note from the bar chart (in Figure \ref{fig:spca_cs3}) that though the difference between the subspace dependence metric values at high SNR values is very small, sPCA has better estimates in almost all runs compared to PCA. The superior performance of sPCA can be attributed the idea of sub-selection of variables. 

With repeated trials, we observed that PCA performs better than sPCA at low SNR values only when there exists repeated or sub-structured  equations in the process information. This can be attributed to identifying all linear relationships at once when the variables present in the linear relationship are same. We improve the performance of sPCA algorithm with appropriate modification in the next subsection. 


\section{Constraint Structural PCA}
\label{sec:cspca}
Structural PCA performed better than PCA when the structural information of the network is known but it can be further improved as discussed in this section. The approach of sPCA algorithm is estimating each linear relation corresponding to a structure separately as seen in Section \ref{sec:SPCA}. All these linear relations were estimated independently in a sequential manner. The key idea in this section for improvising sPCA algorithm is to utilize the information derived up to $(i-1)^\text{th}$ row of the model for estimating the $i^\text{th}$ row of the constraint matrix. 

To utilize the information from first to $(i-1)^\text{th}$ row of the constraint matrix, we present an algorithm termed as Constraint PCA (cPCA) in Appendix section \ref{sec:cpca}. The cPCA algorithm also shows improvement over naive PCA when one or more true equations information is known (or obtained). A detailed discussion and illustrative examples can be seen in Appendix. In this section, we propose a combination of cPCA and sPCA algorithms, termed as CSPCA, which shows improvement over sPCA. 


This combined algorithm can be utilized in presence of repeated equations (i.e. two or more equations involving the same set of variables) or sub-structured equations (i.e. the variables set involved in an equation is a subset of the variables set involved in another equation) in the structural information that is available. It is interesting to note that in the absence of repeated or sub-structured equations in the structural information provided this algorithm results same as sPCA. The pseudo code of the algorithms is as follows:

\begin{enumerate}
    \item Arrange the equations in ascending order of the variables that are involved in individual equations.
    \item For all equations 1 to N, identify the variables set $\phi_i$ that are active in each equation. So, $\phi_i = \{j \quad | \quad  \mathbf{A}(i,j) \neq 0  \}$
    \item Now for each equation i, identify the equations (j from 1 to $i-1$) such that $\phi_j$ is a subset of $\phi_i$ and store the sub-structured equations indices set $\psi_i$. So this can be formally stated as, $\psi_i = \{ i \quad | \quad \phi_i \subseteq \phi_j \quad \forall \quad  j = \{1,2,\ldots,i-1 \} \}$
    \item Now for each equation i, if the sub-structured equations indices set $\psi_i$ is empty then label the equation as ``S'' else ``C''. This means $\text{label}_i = \{ S: |\psi_i| = 0,\quad  \text{else} \quad C\}$
    \item Now for all the equations that are labelled as ``S'' estimate the equations using sPCA by using structural information of individual equations.
    \item Now for all the equations that are labelled as ``C'' estimate the equations using cPCA, assuming the estimated equations set in $\psi_i$ as known equations. 
    \item Rearrange the equations in the given order and report the final estimated A 
\end{enumerate}

Steps 1-4 in the above algorithm are performed to detect the constraints which could be identified using sPCA and CSPCA. For the case study described in section \ref{sec:spca_ex3}, steps 1-4 are performed and is summarized in Table \ref{tab:cspca_ex1_3}

\begin{table}[thpb]
		\caption{ For equation \ref{eq:spca_ex1}  \label{tab:cspca}}
			\label{tab:cspca_ex1_3}
	\vspace{-0.25cm}
	\centering
	\begin{tabular}{*5c}
	\hline
Rearranged index          &Equation& Variable set $\phi_i$  & sub-structured equations  $(\psi_i)$  & Label \\\hline
1 & $[1,0,1,0,0,0]$          & \{1,3\}   & \{\} & S      \\  \hline
2 & $[1,1,1,1,0,0]$          & \{1,2,3,4\}   & \{1\} & C      \\  \hline
3 & $[1,1,1,1,0,0]$          & \{1,2,3,4\}   & \{1,2\} & C      \\  \hline
4 & $[1,1,1,1,0,1]$          & \{1,2,3,4,6\}   & \{1,2,3\} & C      \\  \hline
				\end{tabular}	
	\end{table}

\begin{figure}[H]
   	\centering
   	\includegraphics[width=0.65\linewidth]{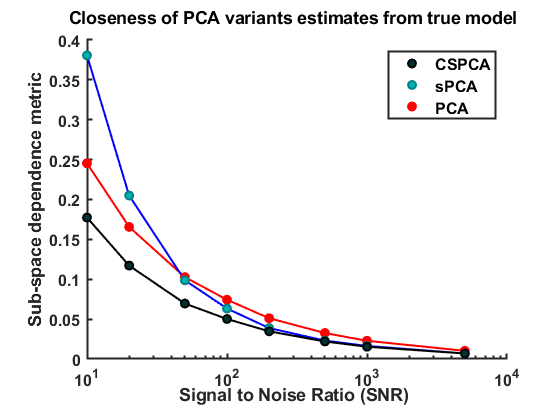}
   	\caption{Comparison of PCA and variants}
   	\label{fig:cspca_ex1}
  \end{figure}

Steps 5-7 are aplied and the results are summarized in Figure \ref{fig:cspca_ex1} for different algorithms. 

 From the above plot, it can be clearly inferred that CSPCA improves the accuracy of sPCA algorithm and outperforms PCA even at low SNR values unlike the case of sPCA.
 
\subsection{ECC Case-study}
\begin{figure}[H]
   	\centering
   	\includegraphics[width=0.65\linewidth]{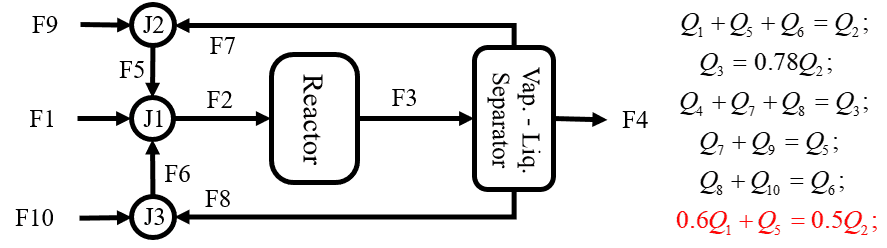}
   	\caption{Flow network of simplified ECC benchmark case study}
   	\label{fig:ecc_case}
\end{figure}

This system is a simplified version of Eastman Chemical Company benchmark case study to test process control and testing methods \cite{paper:downs1993plant}. It involves 10 flows and 6 flow constraints, hence the data is generated by varying 4 flows (F1, F5, F7 and F8) for 1000 time samples. F1 and F2 are mixed streams of reactants A and B with different compositions, F9 and F10 are pure streams of reactant A and B respectively. F3 is a product stream with excess reactants A and B, which are separated using a separator. F4 is a pure product stream, where as F9 and F10 are recycle streams of components A and B. The flow network along with the flow constraints can be observed from Figure \ref{fig:ecc_case}. 

\begin{figure}[H]
	\centering
	\begin{minipage}{.45\textwidth}
		\centering
		\includegraphics[width=0.99\linewidth]{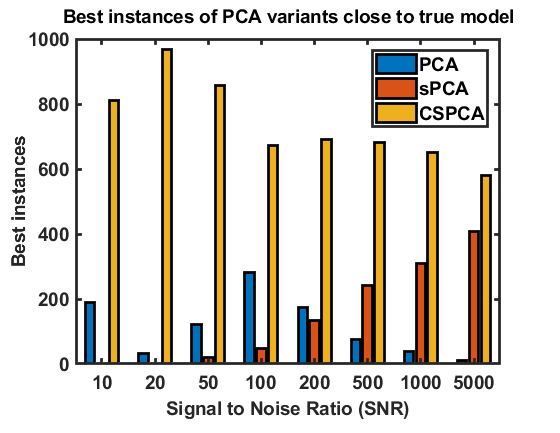}
   	\caption{Frequency of best instances}
   	\label{fig:ecc_pcavsspca2}
	\end{minipage}
	\begin{minipage}{.5\textwidth}
	\centering
	\includegraphics[width=0.96\linewidth]{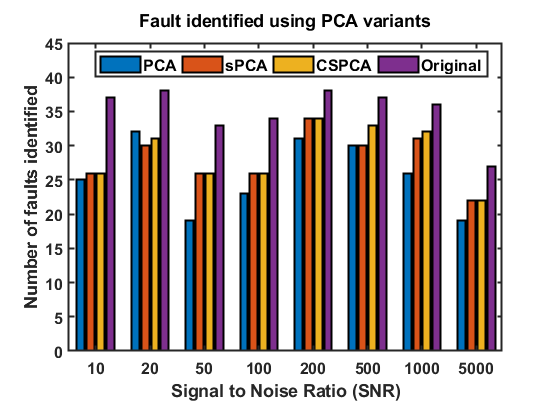}
   	\caption{Comparison of PCA variants for fault detection}
   	\label{fig:ecc_faultdetect}
	\end{minipage}
\end{figure}

The last flow constraint is a material balance constraint of component A at J1. Assuming the structure of the process is known, flow constraint matrix is estimated for 1000 runs of MC simulations using PCA, sPCA and CSPCA for different SNRs. The results by proposed approaches and PCA are presented in Figure \ref{fig:ecc_pcavsspca2}. 

The flow constraint matrices constructed using different algorithms tested to identify the faults in the flow rates of all flows. For illustration, if the flow rates at particular time violates the constraint matrices (sum of the residuals) with in a tolerance limit then the sample considered to be faulty. For different SNR values ( 10, 20, 50, 100, 200, 500, 1000 and 5000), randomly 50 noise added data samples are selected and in each sample one of the variable is randomly modified to make the sample faulty. The flow constraint matrices obtained for the 1000 runs of MC simulations for each SNR value are averaged and considered as final set of flow constraints. The final set of flow constraints obtained using proposed approaches have been tested to identify the faults with a tolerance limit as 1. The number of original faults, which are obtained using the original constraint matrix for the same tolerance are reported along with the number of faults identified using proposed approaches. It can be observed from the table that CSPCA performing better than sPCA, which is superior to PCA.

\section{Conclusion}
\label{sec:conc}
In this study we have formulated model identification schemes, of process models with known structure. To the best of our knowledge, this is the first time such a scheme has been proposed. Implementation of the techniques in the synthetic and real data case-studies have led to improvement over conventional PCA. 

We also proposed the model identification algorithm for the case when few of the linear relations are known apriori. This was termed as constrained PCA. We proposed the combination of cPCA and sPCA which provided further improvement in performance as compared to vanilla PCA and sPCA. The key idea in the integration of two algorithms was to use the information provided by previously estimated linear relations for estimating the further relations. We have also provided general guidelines about the applicability of the combined algorithm.  

Convergence analysis and proposal of highly-scalable version of proposed algorithm is preserved for future work. Another direction of study is identification of constraint matrix structure, which was assumed to be known in this work. 
\section*{Acknowledgment}
We would like to thank Robert Bosch Centre for Data Science and Artificial Intelligence for providing computational facilities.

\appendix
\section*{Appendix}

\section{PCA for Model Identification}
\label{sec:pca_MI}
PCA or total least squares method can be formulated as an optimization problem described below to obtain model parameters. 
\begin{subequations}
	\begin{align}
	\begin{split}
	\min_{\mathbf{A}, \mathbf{x}(t)} & \sum_{t=1}^{N}(\mathbf{y}(t) - \mathbf{x}(t))^\top(\mathbf{y}(t) - \mathbf{x}(t))
	\end{split} \\
	\begin{split} 
	\text{subject to}  & \qquad \mathbf{Ax}(t) = \mathbf{0}_{m \times 1}, \; t = 1, \cdots, N \label{eq:constraint_1}
	\end{split} \\
	\begin{split}
	& \qquad \mathbf{A}\mathbf{A}^\top = \mathbf{I}_{m \times m}
	\label{eq:constraint_2}
	\end{split}
	\end{align}
	\label{eq:objfn_pca}
\end{subequations}
where, $\mathbf{A}$ is referred as the model. It is well known that PCA algorithm utilizes the eigenvalue analysis or equivalently singular value decomposition (SVD) to solve the above optimization problem \cite{jolliffe2002principal, paper:rao1964}. Please note that it is assumed that the number of relations which is the row dimension of $\mathbf{A}$ is known in this work. So, we briefly discuss the utilization of novel eigenvalue decomposition for deriving the model parameters.  

The sample covariance matrix of $\mathbf{Y}$ is defined as 
\begin{align}
\mathbf{S_y} = \frac{1}{N} \mathbf{Y Y^\top} \qquad \mathbf{S_y} \in \mathbb{R}^{n \times n}
\label{eq:sampl_cov}
\end{align}

The eigenvalue decomposition of  sample covariance matrix $\mathbf{S_y} $ is stated as follows:
\begin{align}
\mathbf{S_y U} = \mathbf{ U \Lambda}, \qquad \mathbf{U} \in \mathbb{R}^{n \times n}, \quad \mathbf{\Lambda} \in \mathbb{R}^{n \times n}
\label{eq:eig_decom}
\end{align} 
where $\mathbf{\Lambda}$ is a diagonal matrix containing the eigenvalues and $\mathbf{U}$ consists of the eigenvectors corresponding to those eigenvalues. 

If the noise-free measurements ($\mathbf{X}$ in \ref{eq:N_measr}) are accessible, the constraint model can be derived from the eigenvectors corresponding to zero eigenvalues. This can be intuitively seen by eigenvalue analysis for the covariance matrix of noise-free measurements \cite{paper:rao1964}. 
\begin{align}
\mathbf{S_x U^{\star}} &= \mathbf{ U^{\star} \Lambda^{\star}}, \qquad \qquad \mathbf{S_x} =  \frac{1}{N} \mathbf{X X^\top} \\
\mathbf{S_x U^{\star}_{0}} &= \mathbf{ U^{\star}_{0}} \mathbf{ 0}_{m \times m} = \mathbf{ 0}_{n \times m}, \quad \mathbf{ U^{\star}_{0}} \in \mathbb{R}^{n \times m}  \\
\mathbf{ A_0} & = \mathbf{ U^{\star \top}_{0} }
\end{align}
where, the columns of  $\mathbf{ U^{\star}_{0}}$ contains the eigenvectors corresponding to zero eigenvalues. For the noisy measurements in \eqref{eq:eig_decom}, the eigenvectors corresponding to ``small'' eigenvalues are chosen. For the homoskedastic  case, it can be proved that few of the ``small'' eigenvalues are equal to each other asymptotically and provide an estimate for noise variance in each $n$ variables. It should be noted that PCA provides a set of orthogonal eigenvectors which is a basis for the constraint matrix.


It can be easily proved that PCA provides the total least squares (TLS) solution \citep{jolliffe2002principal} but doesn't grant the freedom to include any available knowledge of process in its formulation. PCA derives the most optimal decomposition based on statistical assumptions without incorporating any process information. Ignoring the underlying network structure leads to minimum cost function value of PCA in \eqref{eq:objfn_pca} but may drive us away from the true process. On the other hand, reformulating  the optimization problem with the inclusion of a priori knowledge as constraints will lead us to a solution closer to true process. Similar approach is adopted in sparse PCA \cite{paper:sparse_pcaTibshirani}, dictionary learning \cite{paper:structured_SPCAjenatton2010}, regularization approaches \cite{paper:pca_lassojolliffe2003,paper:pca_penalized_witten2009} to derive estimates of improved qualities. 



In this section, we briefly discussed PCA and acquired the required background to understand the proposed algorithms in later sections. In the next section, we discuss the approach to utilize the information about a set of linear relations to derive the full constraint matrix / model. 


\section{Model Identification with partially known constraint matrix (cPCA)} 
\label{sec:cpca}
In this section, we assume availability of few linear relationships among $n$ variables. Basically, it is presumed that few rows of the constraint matrix, $\mathbf{ A_0}$ in \eqref{eq:noise_free_mdl} are available. It should be noted that all the linear relationships are not assumed to be known but instead only few of them are available. 

We propose an algorithm termed as constrained principal component analysis (cPCA) to utilize the partially known information of constraint matrix. A simple case-study is considered to illustrate the key idea and assumptions. 

The optimization problem for the partially known constraint matrix can be formally stated below:
\begin{subequations}
	\begin{align}
	\begin{split}
	\min_{\mathbf{A}, \mathbf{x}(t)} & \sum_{t=1}^{N}(\mathbf{y}(t) - \mathbf{x}(t))^\top(\mathbf{y}(t) - \mathbf{x}(t))
	\end{split} \\
	\begin{split} 
	\text{subject to}  & \qquad \mathbf{A_f}\mathbf{ x}(t) = \mathbf{0}_{m \times 1}, \; t = 1, \cdots, N \label{eq:cpca_constr1}
	\end{split} \\
	\begin{split}
	& \qquad \mathbf{A}\mathbf{A}^\top = \mathbf{I}_{l \times l}
	\label{eq:cpca_constr2}
	\end{split}
	\end{align}
	\label{eq:objfn_cpca}
\end{subequations}
where 
\begin{subequations}
	\begin{align}
	\mathbf{ A_f} = \begin{bmatrix}
	\mathbf{ A_{kn}} \\
	\mathbf{ A}
	\end{bmatrix}
	\label{eq:A_full_def}
	\end{align}
	\begin{align}
	\mathbf{ A_f} \in \mathbb{R}^{m \times n}, \quad \mathbf{ A_{kn}} \in \mathbb{R}^{(m-l) \times n}, \quad \mathbf{ A} \in \mathbb{R}^{l \times n}
	\end{align}
\end{subequations}

It is assumed the $(m-l)$ linear equations are known to user and the rest $l$ are to be estimated. Subscripts $(\cdot)_\mathbf{f}$ and $(\cdot)_\mathbf{kn}$, in $\mathbf{ A_f}$ and $\mathbf{ A_{kn}}$, correspond to full and known constraint matrix respectively.  It should be noted the second constraint in \eqref{eq:cpca_constr2} is imposed only on the unknown segment of full constraint matrix to obtain a unique subspace up to a rotation. 

Reconsider a simple flow mixing network example shown in Figure \ref{fig:fl-mix}. 
For this case-study, we assume to have a priori knowledge of the linear relation generated by flow balance on node 1. Therefore,   
\begin{align}
\mathbf{ A_{kn}} = \begin{bmatrix}
1 & -1 & 0 & 0 & 1
\end{bmatrix}
\label{eq:flmx_knwneq}
\end{align} 

One of the naive approaches would be applying PCA without utilizing the knowledge about known linear relation. 
Eigenvalue decomposition of the sample covariance matrix defined in \eqref{eq:sampl_cov} is adopted to obtain the constraint matrix estimate by PCA, denoted by $\hat{\mathbf{ A}}_{pca}$. The eigenvectors corresponding to three smallest eigenvalues provide $\hat{\mathbf{ A}}_{pca}$
\begin{align}
\hat{\mathbf{ A}}_{pca} = \begin{bmatrix}
-0.23     &    -0.49  &        0.02   &       0.70   &       0.46\\
0.12      &    0.49   &      -0.79    &      0.20    &      0.30\\
0.74      &   -0.39   &      -0.05    &     -0.32    &      0.44
\end{bmatrix}
\label{eq:pca_flwmix}
\end{align}
It may be argued intuitively that applying PCA directly in the above case by ignoring the available information will drive the user away from true system configuration. This will be later used for comparison to the proposed method.   

We proceed to discuss the proposed algorithm termed as constrained principal component analysis (cPCA). The objective of this algorithm is to utilize the available information and estimate only the unknown part of constraint matrix as formulated in \eqref{eq:objfn_cpca}. 

For any general known part of constraint matrix, $\mathbf{ A_{kn}} \in \mathbb{R}^{(m-l) \times n}$
\begin{align}
\mathbf{ A_{kn} y}(t) = \mathbf{ A_{kn} x}(t) + \mathbf{ A_{kn} e}(t) = \mathbf{ A_{kn} e}(t) \qquad  \forall \quad t
\end{align}
For a collection of $N$ measurements defined in \eqref{eq:N_measr}, the above may be restated as,
\begin{align}
\mathbf{ A_{kn} Y} = \mathbf{ A_{kn} X} + \mathbf{ A_{kn} E} = \mathbf{ A_{kn} E} \qquad 
\end{align}
To estimate a basis for the rest of linear relations, we attempt to work with data projected on to null space of  $\mathbf{ A_{kn}}$. This can be mathematically stated as,
\begin{align}
\mathbf{ A^{\perp}_{kn}} \mathbf{ X_p} = \mathbf{X}, \qquad \mathbf{ A^{\perp}_{kn}} \in \mathbb{R}^{n \times (n-m + l)}, \quad \mathbf{ X_p}  \in \mathbb{R}^{(n-m+l) \times N}
\label{eq:X_pdef}
\end{align}
where $	\mathbf{ A^{\perp}_{kn}}$ can be viewed as a matrix containing the basis vectors for the null space of $\mathbf{ A_{kn}}$. As the noise-free measurements are not available, \eqref{eq:X_pdef} is restated as, 
\begin{align}
\mathbf{ A^{\perp}_{kn}} \mathbf{ X_p} = \mathbf{Y} -\mathbf{E} 
\end{align}
It should be noted that estimating $\mathbf{ X_p}$ given $\mathbf{ A^{\perp}_{kn}}$ and $\mathbf{ Y}$ leads to overdetermined set of equations as there are $n$ equations for each set of the $(n-m+l)$ variables in columns of $\mathbf{ X_p}$.  This leads to a total of $N \times n$ equations in $N(n-m+l)$ variables. An estimate of the projected data on null space of $\mathbf{ A_{kn}}$ can thus be obtained in least squares sense. 
\begin{align}
\mathbf{ \hat{Y}_p} =  (\mathbf{ A^{\perp}_{kn}})^{\dagger} \mathbf{Y} = (\mathbf{ A^{\perp}_{kn}}^\top \mathbf{ A^{\perp}_{kn}})^{-1} \mathbf{ A^{\perp}_{kn}}^\top \mathbf{Y} 
\label{eq:Xproj_est}
\end{align}
where $\mathbf{ \hat{Y}_p}$ denotes an estimate of $\mathbf{ X_p}$ and $(\mathbf{ A^{\perp}_{kn}})^{\dagger}$ denotes the pseudo-inverse of $\mathbf{ A^{\perp}_{kn}}$.

The unknown part of the constraint matrix estimate, denoted by $\mathbf{ A}$ in full constraint matrix  $\mathbf{ A_f}$ presented in \eqref{eq:A_full_def} can be estimated by applying PCA on projected data $\mathbf{ \hat{Y}_p}$ shown in \eqref{eq:Xproj_est}. The sample covariance matrix of projected data can be defined similar to \eqref{eq:sampl_cov},
\begin{align}
\mathbf{S_{y_p}} = \frac{1}{N} \mathbf{ \hat{Y}_p}\mathbf{ \hat{Y}_p}^\top, \qquad \mathbf{S_{y_p}} \in \mathbb{R}^{(n-m+l) \times (n-m+l)} 
\label{eq:proj_cov}
\end{align}
The eigenvalue decomposition of $\mathbf{S_{y_p}} $, as defined in section \ref{sec:pca_MI}, can be written as,
\begin{align}
\mathbf{S_{y_p}} \mathbf{  U_p} = \mathbf{  U_p \Lambda_p}
\end{align}
The eigenvectors corresponding to $l$ smallest eigenvalues in $\mathbf{  \Lambda_p}$, call it $\mathbf{ A_p}$ provides a basis for the constraint matrix of data in projected space. It should be noted that the original data in $n$ - dimensional space was projected in lower $(n-m+l)$ - dimensional space to estimate the $l$ linear relations. 
\begin{align}
\mathbf{  \hat{A}_p} = \left(\mathbf{  U_p}(:,(n-m-1):(n-m+p)\right)^\top, \qquad \mathbf{\hat{A}_p} \in \mathbb{R}^{l \times (n-m+l)}
\end{align}
using the above with \eqref{eq:X_pdef} and \eqref{eq:Xproj_est}, the following can be stated
\begin{subequations}
	\begin{align}
	\mathbf{  \hat{A}_p X_p} &= \mathbf{ 0}_{l \times N}  \\
	\mathbf{  \hat{A}_p}	(\mathbf{ A^{\perp}_{kn}})^{\dagger} \mathbf{  X} & = \mathbf{ 0}_{l \times N} \Longrightarrow \mathbf{ A X}  = \mathbf{ 0}_{l \times N}
	\label{eq:scal_constrmat}
	\end{align}
\end{subequations}
So, the constraint for original n-dimensional space can be obtained from reduced dimensional space by using 
\begin{align}
\mathbf{ \hat{ A}} =  \mathbf{  \hat{A}_p}	(\mathbf{ A^{\perp}_{kn}})^{\dagger} = \mathbf{  \hat{A}_p} (\mathbf{ A^{\perp}_{kn}}^\top \mathbf{ A^{\perp}_{kn}})^{-1} \mathbf{ A^{\perp}_{kn}}^\top, \qquad \mathbf{  A} \in \mathbb{R}^{l \times n}
\label{eq:constrmat_reddim}
\end{align}
The full constraint matrix can be obtained as stated in \eqref{eq:A_full_def}. 

Revisiting the flow-mixing case study of $5$ variables, the full constraint matrix obtained is stated below. Please note that $\mathbf{  A_{kn}}$ is specified in \eqref{eq:flmx_knwneq}.
\begin{align}
\mathbf{  A_{cpca}} = \begin{bmatrix}
\mathbf{  A_{kn}} \\
\mathbf{  \hat{A}}
\end{bmatrix} =
\begin{bmatrix}
1        & -1          &   0           &  0        &  1 \\
-0.53   &      -0.34     &     0.14     &     0.74     &     0.19\\
-0.07   &      -0.42     &     0.77     &    -0.30     &    -0.36
\end{bmatrix}
\label{eq:cpca_flowmix_est}
\end{align}

The true constraint matrix specified in \eqref{eq:flmix_trueA0} is used to evaluate the accuracy of estimates obtained by PCA and cPCA specified in \eqref{eq:pca_flwmix}, \eqref{eq:cpca_flowmix_est}. The subspace metric defined in \eqref{eq:subspace_metr} is used to compare the estimates:
\begin{align}
\theta_{PCA} = 0.1293,  \qquad \theta_{cPCA} = 0.0747
\label{eq:fl-mix_est_pcancpca}
\end{align}
It may be easily inferred that the proposed algorithm cPCA is outperforming PCA using the subspace dependence metric. This simple case-study with synthetic data was presented for the ease of understanding the notations and demonstrating the key idea of cPCA. 


The novel contribution of this work is to wisely utilize the available information about a subset of linear relations and transforming the original problem stated in \eqref{eq:objfn_cpca} to PCA friendly framework. This rewarding step provides us the freedom to include the prior available information and also the ease of implementation through analytical solution by PCA. Basically, this is performed in two steps.  The first one is projecting the data in null space of known linear relation and the second step is applying PCA in the reduced space. Finally, the obtained solution is transformed back from reduced to original space. We close this section with summarizing the algorithm in Table \ref{Tab:cpca_algo}. We show the efficacy of proposed algorithm over PCA on another multivariable case-study in the next subsection. 




\begin{table}[H]

	\caption{Constrained PCA (cPCA) Algorithm }
	\label{Tab:cpca_algo}
	\hrulefill
	\begin{enumerate}
		\item Obtain the null space $\mathbf{  A^{\perp}_{kn}}$ for given set $(m-l)$ of linear relations among $n$ variables.
		\item Obtain the projection of data $\mathbf{ \hat{Y}_p}$, onto the null space $\mathbf{  A^{\perp}_{kn}}$ using \eqref{eq:Xproj_est}. 
		\item Apply PCA on the lower dimension projected data  $\mathbf{ \hat{Y}_p}$ to obtain $\mathbf{  \hat{A}_p }$.
		\item Transform the estimated $\mathbf{  \hat{A}_p }$ in previous step to original subspace using \eqref{eq:constrmat_reddim}. The full constraint matrix can be constructed using \eqref{eq:A_full_def}.
	\end{enumerate}
\hrulefill
\end{table}
\bibliography{refs}

\end{document}